\documentclass[runningheads]{llncs}
\usepackage[T1]{fontenc}
\usepackage{graphicx}
\usepackage{amsmath}
\usepackage{booktabs}
\usepackage{subcaption}
\usepackage{algorithm}
\usepackage{algorithmic}
\usepackage{hyperref}
\usepackage{orcidlink}
\usepackage{color}

\begin{document}
\title{Diversity Preservation Recovers Transcription Factor Motifs Across Challenging Confounder Landscapes}

\titlerunning{Diversity Preservation Recovers TF Motifs Across Challenging Landscapes}

\author{Alejandro Medina \orcidlink{0009-0005-7618-4947} \and
Mary Lauren Benton \orcidlink{0000-0002-5485-1041}}
\authorrunning{A. Medina and M.L. Benton}

\institute{Baylor University, Waco TX 76798, USA\\
\email{\{alejandro\_medina1, marylauren\_benton\}@baylor.edu}}

\maketitle              
\begin{abstract}
Motif discovery is a process where DNA sequences are searched for the short patterns, called motifs, that transcription factor (TF) proteins bind. However, the motif lengths are often unknown and motif attribution often competes with a near-identical motif of a related protein. In this work, we pair a motif genotype whose length emerges from an evolvable gate with MAP-Elites archives organized by three descriptor pairs, and validate every motif against the full HOCOMOCO database. On five liver ChIP-seq datasets spanning distinct confounder landscapes, a $(\mu+\lambda)$ baseline differing only in survivor selection collapses onto a single motif and never ranks the target first across three datasets, while every MAP-Elites archive does so in at least 13 of 30 seeds on every dataset. The archives recover the target signal as reliably as the motif discovery tool MEME on four datasets and more robustly when motif width is constrained. Neither method ranks the target above its closest relatives on HNF4A and JUND over the full width range, but narrow archive motifs rank HNF4A first in up to 43\% of seeds, where MEME misses the motif in all settings. Diversity preservation improves our ability to recover consistent motif signal, although motif annotation remains a challenging problem for all methods we tested. 

\end{abstract}

\section{Introduction}
\label{sec:introduction}
    A transcription factor (TF) is a protein that binds short DNA sequences, called motifs, to regulate genes. Chromatin immunoprecipitation followed by sequencing (ChIP-seq) experiments \cite{Johnson2007} can locate where a TF was bound across the genome in living cells, and motif discovery can then search the sequences of the bound regions for the binding motif(s). Methods that report a single best motif assume that the top-ranked candidate is the correct one. That assumption holds when a TF has one dominant binding preference, but it can fail when the TF shares its binding site with a close relative whose motif is nearly indistinguishable by sequence alone \cite{Pugacheva2015}. We use \emph{confounder} for any protein, such as a paralog (a TF that arose from the duplication of a shared ancestral gene) or an unrelated co-occupying factor, whose motif is similar enough to the target's that a search cannot reliably tell them apart. For example, the paralog CTCFL recognizes a motif nearly identical to CTCF \cite{Pugacheva2015}.
    
    Algorithms for motif identification face two big challenges. The first is that a motif's informative length is not always known in advance and varies across TFs \cite{Aptekmann2022}. The biological term for this quantity is \emph{width}; however, we use the term `length' because we treat each candidate as a sequence, and use width only in reference to tool settings and the gate parameter in Section~\ref{sec:genotype}. Established tools take a width or width range as input and rank motifs by statistical score \cite{Bailey2021,Bailey1994,heinz_simple_2010}, while evolutionary and deep-learning approaches fix length in advance or predict it in a separate step \cite{Alipanahi2015,che_mdga_2005,10.1109/TCBB.2022.3158471}. The second challenge is that the correct motif competes with relatives that score almost as well, so returning a single top-ranked answer can obscure other competitive sequences that were found during the search. The earlier quality-diversity (QD) study of this problem \cite{medina_motif_2026} fixed motif length, used one dataset, and reported only matches to the target factor. 
    
    The approach introduced in \cite{medina_motif_2026} is extended in this paper to address both challenges. We use MAP-Elites (Multi-dimensional Archive of Phenotypic Elites \cite{mouret_illuminating_2015}) over a variable-length genotype whose length emerges from an evolved gate, with archives organized by descriptors grounded in binding geometry. Every archived motif is validated against the full HOCOMOCO database \cite{Vorontsov2023} with the motif-comparison tool TomTom \cite{gupta_quantifying_2007}, which distinguishes a match to the target from a match to a competitor. In evolutionary terms, the fitness is a proxy that cannot separate near-tied optima, which is where diversity preservation should help. This case study asks three questions using five liver ChIP-seq datasets from ENCODE \cite{the_encode_project_consortium_integrated_2012} chosen to span four confounder landscapes: (1) Does diversity preservation recover the target where a matched $(\mu + \lambda)$ elitist baseline does not? (2) Does an established approach, MEME \cite{Bailey1994}, validated by the same pipeline, avoid the same difficulty? (3) What additional information does the archive provide that a ranked list does not?  
    
    We evaluate performance using two distinct success criteria: any-hit and best-hit. A candidate is an \emph{any-hit} if the target appears anywhere among its significant TomTom matches, and a \emph{best-hit} if the target is its closest match. Any-hit asks whether the search recovered the target signal, and best-hit asks whether the target outranks its competitors, which no sequence-only method can guarantee against a nearly identical motif. Only the exact target counts; family and confounder matches are reported but never counted as hits. 

    This paper makes three contributions. First, it shows that diversity preservation is required to recover the target reliably on the four datasets with several close competitors, but is unnecessary for CTCF, where fitness alone separates the target from its competitor. 
    Second, it shows that MAP-Elites matches MEME's recovery on four of five datasets under identical inputs and validation and is more robust across motif length, although neither method ranks the target above its closest relatives for HNF4A and JUND. Finally, it extends earlier fixed-length QD motif discovery \cite{medina_motif_2026} to a variable-length genotype and introduces an evaluation protocol for methods that return sets of motifs, with exact-target any-hit and best-hit criteria and validation against a full reference database.  

\section{Related Work}
\label{sec:related-works}
    Computational motif discovery spans classical statistical search, including MEME \cite{Bailey1994}, HOMER \cite{heinz_simple_2010}, and STREME \cite{Bailey2021}, deep-learning models such as DeepBind \cite{Alipanahi2015}, and evolutionary search \cite{che_mdga_2005,Gonzlezlvarez2014}. Classical tools accept a motif width or a range and return several ranked motifs, evolutionary methods typically fix the length in advance, and deep-learning approaches predict length in a separate step \cite{10.1109/TCBB.2022.3158471}. The closest prior treatment of variable length is a niched Pareto genetic algorithm (GA) that treats length and consensus similarity as competing objectives \cite{Vijayvargiya2011}. None of these keeps structurally different motifs from one search side by side, which is what QD algorithms are designed to do.

    The only prior application of QD to TF motif discovery that we are aware of used MAP-Elites, which keeps the best motif found in each region of a space of motif properties, and showed that the approach is feasible \cite{medina_motif_2026}. It evolved position weight matrices on ChIP-seq data for CTCF and obtained several distinct motif variants with fitness close to that of the best MEME motif. Its limitations motivate this paper. Motif length was fixed at the canonical CTCF length, although this property is unknown for most factors and the archive could not show how quality changes with length. It used one dataset, and CTCF has a single close competitor, which makes it the easiest of our five, so the study could not show whether diversity helps when competition is crowded. Finally, its benchmark compared MEME by fitness under the QD fitness function, which favors the method that optimizes for this quantity, and had no matched baseline without diversity preservation.  

    QD algorithms return a diverse set of high-performing solutions in place of a single optimum \cite{pugh_quality_2016}. MAP-Elites \cite{mouret_illuminating_2015} divides a space of user-chosen behavior descriptors into a grid of cells and keeps the fittest individual found in each. Each iteration samples parents from the occupied cells, varies them, and places every offspring in the cell its descriptors select; it replaces the occupant only if the cell is empty or the offspring is fitter. Selection therefore acts within a cell and not across the population, so solutions that differ in behavior survive even when they are less fit than the best. QD was developed in robotics \cite{cully_robots_2015} and has been applied to protein design \cite{boige_gradient-informed_2023} and to k-mer feature selection for classifying SARS-CoV-2 sequences \cite{Sambhe2021}, but neither of these biological applications addresses DNA regulatory motifs.

    A method that returns a set of results raises the question of how success is scored. Multimodal optimization scores the fraction of known optima recovered within a tolerance \cite{Li2017}, and QD evaluation depends on the archive and its discretization \cite{Kent2022}. Published evaluations may score the best of several reported motifs against a reference motif for the factor or a family member \cite{Bailey2021}, or accept any known motif of a curated factor group and keep additional enriched motifs, in which case a discovered motif can outrank the known motif of the factor \cite{Kheradpour2013}. Mixed attribution is expected because related DNA-binding domains prefer similar motifs \cite{Pabo1992,Weirauch2014} and only about one fifth of footprints can be unambiguously assigned to a single TF \cite{Vierstra2020}. We therefore report two success criteria and identify the competitors discovered by our approach. 

\section{Method}
\label{sec:method}

    \subsection{Task and Data}
    \label{sec:task-and-data}
    We use ChIP-seq peak calls for five TFs in human liver from the ENCODE Project \cite{the_encode_project_consortium_integrated_2012} (Table~\ref{tab:data}). The JUND, YY1, HNF4A, and RXRA samples come from a single female donor, and CTCF pools two donors. The datasets span four confounder landscapes, which are the problem types of interest: (i) one paralog with a near-identical motif that is expressed mainly in germ cells, so in liver it competes as a look-alike in the reference database rather than at the peaks (CTCF \cite{Pugacheva2015}), (ii) a paralog plus an unrelated protein that binds the same regions (YY1, a core-promoter factor \cite{Xi2007}, whose paralog YY2 and promoter partner TAF1 both appear among its archive matches in Figure~\ref{fig:composite}), (iii) many same-family relatives with near-identical motifs (JUND \cite{Fonseca2019}), and (iv) a crowded TF family built around one shared DNA element (RXRA and HNF4A \cite{Fang2012}). We chose these datasets for this diversity of landscapes rather than ease of recovery. 

    For each peak we extract a 200 base pair (bp) window centered on the peak summit, and assign it the peak's narrowPeak signal value. A window that overlaps several peaks is assigned the signal of the largest overlap. The background set matches the foreground in size and length using regions outside every peak.

    \begin{table}[t]
    \centering
    \caption{Datasets and expected competitors used for labeling. Peaks are 200 bp summit-centered windows.}
    \label{tab:data}
    \begin{tabular}{@{}lllll@{}}
    \toprule
    TF & Class & ENCODE file & Expected competitors & Peaks \\
    \midrule
        CTCF & zinc finger & ENCFF143HEE & CTCFL & 38{,}935 \\
        JUND & bZIP & ENCFF420PED & JUN, JUNB, FOS, ATF, BATF & 22{,}701 \\
        YY1 & zinc finger & ENCFF838VFX & YY2 (TAF1 co-binds) & 31{,}893 \\
        HNF4A & nuclear receptor & ENCFF837QHJ & HNF4G, RXRA, NR2F1/2 & 69{,}131 \\
        RXRA & nuclear receptor & ENCFF572MCI & RXRB/G, RAR & 57{,}430 \\
    \bottomrule
    \end{tabular}
    \end{table}

    \subsection{Variable-Length Motif Genotype}
    \label{sec:genotype}
    A motif's length is not known in advance, so the representation must allow length to change without making individuals incomparable. To accomplish this, we give every individual the same fixed array and let a smooth gate decide how much of it is expressed, so that length is a continuous trait varied by the same operator as every other trait. Each individual is a real-valued vector with four base logits at each of $L_{max}=30$ positions, following the standard position weight matrix (PWM) \cite{stormo_dna_2000}, plus a gate center $c$ and a gate width $w$, for $122$ values. A softmax over the four logits gives a distribution over A, C, G and T at each position. The gate is the product of a rising and a falling logistic edge,
    \begin{equation}
    g_i = \sigma\!\left(\frac{i-(c-w/2)}{\tau}\right)\sigma\!\left(\frac{(c+w/2)-i}{\tau}\right),
    \end{equation}
    with $\sigma$ the logistic function and $\tau = 1$. A position's log-odds contribution is multiplied by $g_i$, so positions outside the window contribute nothing. The width is kept between $5$ and $30$, which covers the $5$ to $20$ bp typical of binding sites \cite{Aptekmann2022}, and the window inside the array. The effective length $\ell(g) = \Sigma_ig_i$ is approximately $w$, so length is a property of each individual, and individuals of different lengths compete in one archive.  

    \subsection{Fitness Function}
    \label{sec:fitness-function}
    The fitness must reward patterns enriched in strongly bound sequence without consulting any motif database, so that validation against HOCOMOCO remains independent. It keeps the structure of the earlier study's fitness~\cite{medina_motif_2026}, a log-odds discrimination of foreground from background sequence, and adds three things for variable-length motifs: the gate, a length-dependent null, and an information term. Each genotype $g$ is scored as
    \begin{equation}
    F(g) = S(g) + \beta\, I(g), \qquad \beta = 0.4.
    \end{equation}
    The match score $S(g)$ is computed in three steps. The gated motif is converted to log-odds against the background composition of the dataset, and the best placement on either strand is kept for each window. These scores are divided by the effective length $\ell(g)$ and averaged, after trimming 10\% at each end, over the 20\% of windows with the highest ChIP-seq signal. The result is then standardized against chance scores at the same length, using the mean and standard deviation of 8000 random genotypes per dataset, each fitted as a cubic polynomial of length. This
    standardization removes the dependence of chance scores on length, but it is not a significance test, because evolved genotypes score far above the random genotypes used to fit it. The information term $I(g) = \sum_i g_i(2 - H_i)$ is the gate-weighted sum of per-position information in bits, where $g_i$ is the gate value and $H_i$ is the Shannon entropy of position $i$. It is summed and not averaged, so it outweighs the advantage of very short motifs, which score well simply because fewer positions can disagree with a window. We chose $\beta$ in preliminary runs on CTCF, YY1, and JUND, where smaller values let degenerate short motifs dominate the archive, and fixed it for all five datasets.

    \subsection{Descriptors and MAP-Elites}
    \label{sec:descriptors}
    Each descriptor pair combines effective length with one sequence property of TF binding that the fitness does not control, so each archive holds motifs that score alike and differ in structure. Length is the first axis of every pair because it is the property the genotype makes evolvable and the one that cannot vary with a fixed-length tool. \emph{Composition} (length, GC fraction) and \emph{Degeneracy} (length, positional entropy) reuse two descriptors of the earlier study \cite{medina_motif_2026}, where they formed a single pair. Both are redefined for the gated genotype, with the ranges given below: each position contributes in proportion to its gate value, so only the expressed part of the motif counts and the values are comparable across lengths. Composition is the gate-weighted GC fraction. GC-rich motifs are typical of zinc finger proteins and $F(g)$ has no composition term, so motifs with very different GC content can score alike. Degeneracy is the gate-weighted mean per-position entropy in nats. It is not orthogonal to the fitness, because $I(g)$ is built from the same entropy, so it separates motifs the fitness already rewards from motifs that reach a similar score while staying more permissive at individual positions. \emph{Palindromy} (length, reverse-complement symmetry) is new to this paper. It is the similarity between the position-wise base distributions and the same distributions read in reverse and complemented, computed as one minus half the mean L1 distance, so a perfect reverse-complement palindrome scores 1. TFs that bind as symmetric homodimers often recognize palindromic or near-palindromic motifs \cite{Pabo1992}, and the fitness cannot tell a symmetric preference from an asymmetric alternative that scores the same.

    Each pair defines its own archive. Effective length runs from 1--30 bp, the maximum searched, and the second axis spans its theoretical range (GC fraction 0 to 1, symmetry 0 to 1, entropy 0 to $\ln 4$ nats). Each axis has 25 equal intervals, giving a $25\times25$ archive of 625 cells, and the three archives are run independently. A resolution ablation ($10\times10$, $15\times15$, $15\times25$, $25\times15$, $25\times50$; 10 seeds) showed no significant benefit at any other resolution, so we report $25\times25$. The archive is seeded with $512$ random individuals. Each iteration produces $512$ offspring, each from two parents sampled uniformly from the occupied cells with the iso+line operator \cite{vassiliades_mouret_isoline_2018} as implemented in QDax \cite{chalumeau2024qdax}. The operator adds isotropic Gaussian noise to the first parent plus a displacement toward the second, with $\sigma_{iso} = 0.2$ and $\sigma_{line} = 0.25$ on the logits and $0.5$ and $0.25$ on the gate.

    \subsection{Biological Validation}
    \label{sec:biological-validation}
    Every motif is compared with all 1{,}595 motifs of HOCOMOCO v14 CORE \cite{Vorontsov2023} with TomTom 5.5.9 \cite{gupta_quantifying_2007} using Pearson correlation, a minimum overlap of 5, and an E-value reporting threshold of 1.0. A match is kept if its q-value is at most 0.05. The reporting threshold matters as a match can have $q \le 0.05$ with $E > 0.05$, and in our data such matches are one third to one half of all significant target matches on JUND, RXRA, and YY1, so a stricter E-value cutoff would drop them. Matches are ordered by p-value. Each top match is labeled `target', `family', `confounder', or `other', using the per-dataset competitor lists in Table~\ref{tab:data} and a fixed list of GC-rich zinc-finger proteins whose motifs resemble many generic GC-rich patterns. A motif with no TomTom match at $q\le 0.05$ is labeled `no match', meaning that nothing in HOCOMOCO resembles it at the significance threshold, not that it is a negative result. Where results are summarized by category (Table~\ref{tab:category}), family and confounder matches are pooled, and `other' is reported as a wrong motif.

\section{Experiments}

    \subsection{Baseline}
    \label{sec:baseline}
    To show what diversity preservation adds, we compared MAP-Elites against an elitist evolutionary algorithm with $(\mu+\lambda)$ selection \cite{Beyer2002} that shares its genotype, fitness function, and iso+line variation, and differs only in how survivors are chosen. Parents are drawn uniformly with replacement, and the best $\mu = 625$ of the $(\mu + \lambda)$ pool survive. We set $\lambda = 512$ to match the MAP-Elites batch size, so both methods evaluate the same number of candidates per iteration and end with at most 625 individuals. The baseline starts from $625$ random genotypes independent of MAP-Elites' $512$ and does not use descriptors. All configurations (three descriptor pairs and the baseline) run for $5000$ iterations of $512$ evaluations on each dataset with $30$ seeds in QDax \cite{chalumeau2024qdax}. Successful runs, those with at least one best-hit individual, are compared between each descriptor pair and the baseline with two-sided Fisher's exact tests, Holm-adjusted over the 15 comparisons. This criterion is lenient, since an archive offers many chances at one hit, so Section~\ref{sec:meme-vs-me} repeats the comparison after removing redundant motifs.  

    \subsection{MEME Comparison}
    \label{sec:meme}
    MEME 5.5.9 \cite{bailey_meme_2015} is run on the same foreground and background as the QD runs in its classic mode, with the zero-or-one-occurrence-per-sequence model (zoops), both strands, a zero-order background model built from the background set, and ten motifs per run. We run it once over the archive's full length range (6--30 bp) and once in each of five length windows (6--10, 11--15, 16--20, 21--25, 26--30 bp), which mirrors restricting the archive to a band of its length axis. Every reported motif, including those MEME scores as non-significant, is validated with an identical TomTom command and threshold. MEME is deterministic for a fixed input, so each condition is a single run, reported as a reference against the distribution over 30 QD seeds rather than tested statistically. We use the classic objective because, like our fitness function, it sees only the background composition and not the background sequences; the discriminative mode is left for future comparison. 

    To compare equal candidate budgets we take ten motifs from each archive, over the same length range or within the same window. An archive's ten fittest individuals are usually near-copies of one motif from neighboring cells, and each copy is another chance at a match. We therefore follow the fitness ranking but skip any individual whose motif shares a core with one already chosen, meaning their informative columns (at least 0.5 bits, ends trimmed) correlate at Pearson $r \ge 0.75$ with the best ungapped alignment on either strand, over at least 70\% of the shorter core. We also skip motifs shorter than 6 bp, which MEME cannot return under these settings; including them does not change the any-hit or best-hit rate. The same rule finds 5--10 distinct cores among each MEME run's ten motifs. Because we set the threshold, we repeat the analysis at 0.6 and 0.9. A seed is successful under a criterion if at least one of its deduplicated motifs meets it. Because the target can appear deep in a long list of matches to a generic motif, we also report a strict any-hit requiring the target among a motif's five closest matches; the cutoff of five is arbitrary and serves only to flag such buried hits. 

\section{Results}

\subsection{QD Search Against the Elitist Baseline}
\label{sec:diversity-effect}
The fitness of the baseline’s best individual reaches the level of MAP-Elites for every seed in the CTCF dataset, but in only 1 to 16 of 30 seeds on the other four datasets. This makes its mean best significantly lower on HNF4A, YY1 and JUND and it is far more variable between seeds (max SD of 2.22, compared to max SD of 0.49 for MAP-Elites; Table~\ref{tab:fitness}). On RXRA the difference is not significant because the baseline's seeds split between the higher peak and a lower one. The baseline's mean fitness is higher than every MAP-Elites pair's on every dataset, which is the signature of convergence. Without a mechanism that preserves diversity, elitist selection concentrates the population near one peak, and the recovery results show that the peak is often a confounder. At lengths 6--30 a baseline population holds at most about one distinct motif core, and on HNF4A and JUND nearly half of seeds end with only motifs shorter than 6 bp. MAP-Elites archives stay broad, with 83.5--88.0\% of cells filled.  

    \begin{table}[t]
    \centering
    \caption{Mean fitness per archive or population and best fitness (each seed's fittest individual, mean $\pm$ SD over 30 seeds). Every MAP-Elites Best differs from the baseline's (Mann--Whitney U, Holm-adjusted $p<0.05$) except on RXRA.}\label{tab:fitness}
    \begin{tabular}{lcccccccc}
    \toprule
    & \multicolumn{6}{c}{MAP-Elites} & \multicolumn{2}{c}{Baseline}\\
    \cmidrule(lr){2-7}\cmidrule(lr){8-9}
    & \multicolumn{2}{c}{Composition} & \multicolumn{2}{c}{Palindromy} & \multicolumn{2}{c}{Degeneracy} & \multicolumn{2}{c}{$(\mu+\lambda)$}\\
    Dataset & Mean & Best & Mean & Best & Mean & Best & Mean & Best\\
    \midrule
    CTCF  & 12.06 & 30.65$\pm$0.19 & 20.77 & 30.78$\pm$0.17 & 14.34 & 30.61$\pm$0.33 & 30.91 & \textbf{31.00}$\pm$0.19\\
    HNF4A & 6.03 & 15.27$\pm$0.14 & 10.59 & \textbf{15.35}$\pm$0.13 & 3.05 & 15.25$\pm$0.12 & 11.58 & 11.61$\pm$2.22\\
    RXRA  & 5.19 & 12.61$\pm$0.15 & 9.59 & \textbf{12.63}$\pm$0.18 & 1.37 & 12.50$\pm$0.24 & 11.92 & 11.96$\pm$1.11\\
    YY1   & 5.66 & 12.43$\pm$0.22 & 9.82 & \textbf{12.44}$\pm$0.16 & 1.62 & 12.10$\pm$0.26 & 11.53 & 11.58$\pm$0.23\\
    JUND  & 4.20 & \textbf{10.73}$\pm$0.40 & 7.66 & 10.56$\pm$0.49 & $-$1.11 & 10.50$\pm$0.42 & 9.69 & 9.71$\pm$0.74\\
    \bottomrule
    \end{tabular}
    \end{table}

\begin{figure}[t]
        \centering
        \includegraphics[width=.85\linewidth]{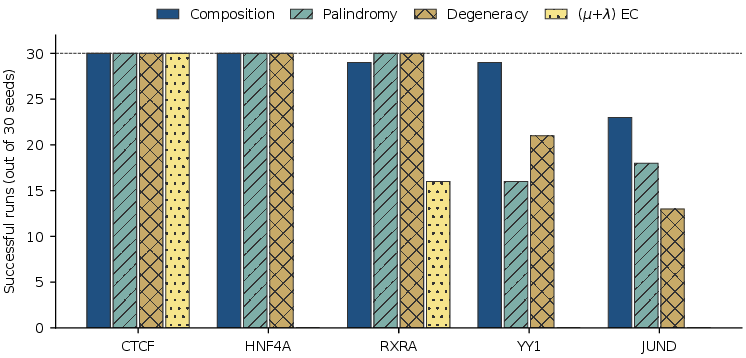}
        \caption{Successful runs (at least one best-hit, out of 30 seeds) by descriptor pair and baseline.}
        \label{fig:successfulruns}
    \end{figure}

The baseline produces no best-hit in any of 30 seeds for HNF4A, YY1, and JUND. For RXRA, 14 of 30 runs show collapse toward a dominant nuclear-receptor confounder, leading to no best-hit in those runs. MAP-Elites produces a best-hit in at least 13 of 30 seeds for every pair on every dataset (Figure~\ref{fig:successfulruns}), and in at least 29 of 30 on CTCF, HNF4A, and RXRA. JUND is the hardest TF, with 23, 18, and 13 successful seeds for Composition, Palindromy, and Degeneracy. Composition is the most reliable pair on YY1 (29 of 30, against 16 and 21). Every pair is more reliable than the baseline on HNF4A, RXRA, YY1, and JUND (Fisher's exact test, Holm-adjusted $p \le 6.8 \times 10^{-4}$). The baseline succeeds in all 30 seeds on CTCF because fitness ranks the target above its single close competitor, suggesting that diversity matters more when the competition is crowded.

\subsection{Fitness Captures Biological Signal but Not Rank}
\label{sec:biological-signal}
Target-hit individuals have the highest mean fitness in every dataset, and the point-biserial correlation between fitness and any-hit status is 0.19 to 0.74 across datasets and descriptor pairs. The correlation with best-hit status stays at or below 0.15 on four of five datasets and reaches 0.38 to 0.65 only on CTCF, because a motif that fits the bound sequences well usually matches the target and its close relatives alike, and the one relative's match is often the stronger one. We quantify this competition as follows. The strongest competitor is the protein with the highest count, and its margin is that count divided by the next protein. A margin near 1 means several competitors are tied, and a large margin means one competitor stands alone. We compute margins over the whole archive and over the top 1\% of motifs by fitness. For CTCF, CTCFL is the strongest competitor, and its margin grows from 9.6x across the archive to 236x among the top 1\% by fitness, where the target is the top match of every motif and CTCFL is the only competitor that appears, so fitness alone separates the target from its rival. HNF4A and RXRA have no dominant paralog at any level, because several nuclear receptors match nearly the same motifs, with the strongest competitor's count within 1.1x to 1.2x of the next. YY1's competitors are its paralog YY2 and a non-homologous partner, TAF1. On JUND the whole-archive competitors are unrelated zinc fingers, and the AP-1 relatives appear only among the fittest individuals. Therefore best-hit alone considers CTCF solved, meaning the target is the top match in every seed, and JUND nearly unsolved, with a best-hit in at most 7\% of seeds, although the target signal was recovered in both cases.

\begin{table}[t]
    \centering
    \caption{Mean fitness by TomTom match category, pooled over descriptor pairs and 30 seeds. Ranges ($^{*}$) span the three descriptor pairs.}\label{tab:category}
    \setlength{\tabcolsep}{5pt}
    \begin{tabular}{lcccc}
    \toprule
    Dataset & Target hit & Family/confound & Wrong motif & No match\\
    \midrule
    CTCF  & 19.59 & 9.28             & $-1.98$ & 7.17\\
    HNF4A & 10.17 & 0.34--2.88$^{*}$ & 0.59    & 6.18\\
    RXRA  & 8.84  & 3.25--4.18$^{*}$ & $-0.89$ & 4.93\\
    YY1   & 10.64 & 9.93             & 3.12    & 6.80\\
    JUND  & 8.87  & 5.26--6.82$^{*}$ & 1.57    & 3.69\\
    \bottomrule
    \end{tabular}
    \end{table}

\subsection{MAP-Elites Against MEME}
\label{sec:meme-vs-me}
Over the full length range, MAP-Elites recovers the target signal as reliably as MEME on four of five datasets (Table~\ref{tab:meme}). Both approaches rank CTCF first every time. For HNF4A and RXRA, every archive recovers the target in all 30 seeds, and for YY1 the target is recovered in 87--97\% of seeds, matching MEME, whose single run recovers the target on every dataset. For JUND, MEME recovers the AP-1 motif in every run, while the archives do so in 47--80\% of seeds, and the target motif is deeper in their match lists (any-hit within the five closest matches in 20--30\% of seeds). The baseline falls below MEME on every dataset except CTCF (0--57\% of seeds), suggesting that diversity preservation is what brings evolutionary search to the level of the classical tool. On RXRA the baseline's best-hit rate falls from 16 of 30 seeds in Figure~\ref{fig:successfulruns} to none in Table~\ref{tab:meme}, because Figure~\ref{fig:successfulruns} counts any individual in the 625-member population, whereas the deduplicated top ten of a converged baseline holds at most one distinct motif. That motif has another nuclear receptor as its top match in 18 of the 24 seeds that have one and no significant match in the other 6, and the remaining 6 seeds have no motif of at least 6 bp.

\begin{table}[t]
    \centering
    \caption{Seeds (\%) with a deduplicated any-hit / best-hit, over 6--30 bp and over the five length windows pooled, against single MEME runs (yes/no, or successful runs of 5). $^{a}$Non-significant based on MEME's E-value. $^{b}$Only one hit has the target among the five closest matches.}\label{tab:meme}
    \setlength{\tabcolsep}{1.9pt}
    \scriptsize
    \begin{tabular}{l ccccc c ccccc}
    \toprule
    & \multicolumn{5}{c}{Full range, 6--30 bp} && \multicolumn{5}{c}{Fixed windows pooled}\\
    \cmidrule(lr){2-6}\cmidrule(lr){8-12}
    Dataset & Comp. & Pal. & Deg. & $(\mu + \lambda)$ & MEME && Comp. & Pal. & Deg. & $(\mu + \lambda)$ & MEME\\
    \midrule
    CTCF & 100/100 & 100/100 & 100/100 & 100/100 & yes/yes && 100/100 & 100/100 & 100/100 & 100/100 & 5/5\\
    HNF4A & 100/0 & 100/0 & 100/3 & 23/0 & yes/no && 100/43 & 100/37 & 100/30 & 23/0 & 5/0\\
    RXRA & 100/0 & 100/3 & 100/13 & 57/0 & yes/no && 100/7 & 100/7 & 100/27 & 57/0 & 4$^{b}$/1\\
    YY1 & 97/13 & 87/0 & 87/3 & 13/0 & yes/no && 97/30 & 97/0 & 87/17 & 13/0 & 1$^{a}$/0\\
    JUND & 80/3 & 60/7 & 47/0 & 0/0 & yes/no && 80/17 & 60/7 & 47/3 & 0/0 & 5/0\\
    \bottomrule
    \end{tabular}
    \end{table}

\begin{figure}[p]
    \centering
    \includegraphics[width=\textwidth]{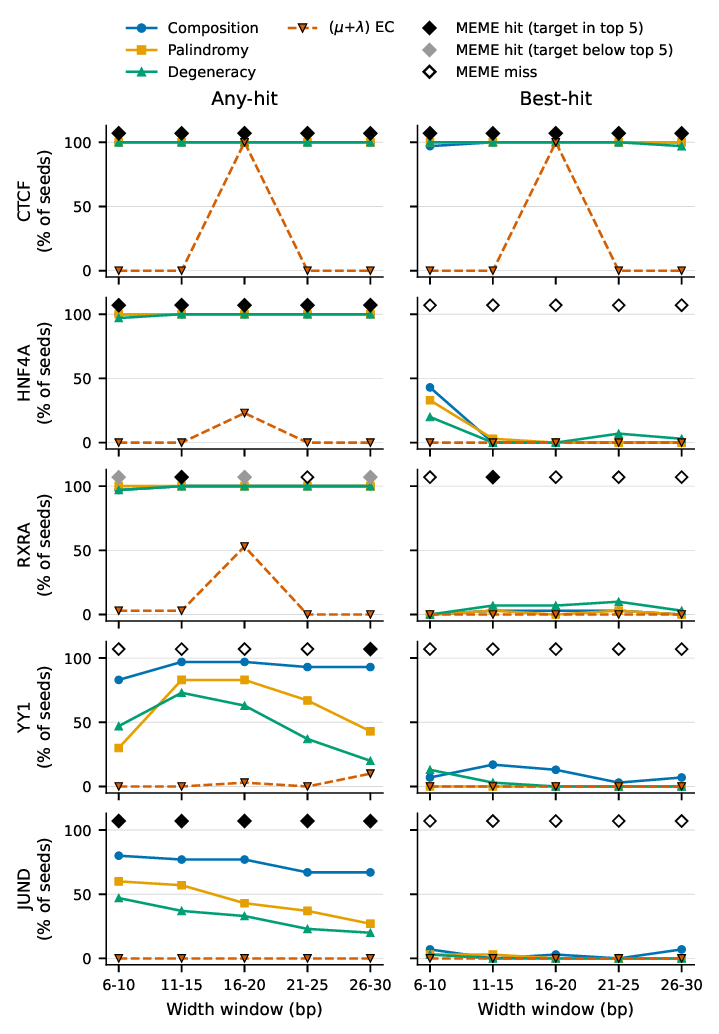}
    \caption{Recovery within each length window, as the percentage of 30 seeds whose deduplicated top ten contains an any-hit (left) or best-hit (right). Diamonds mark the MEME run for the same window.}\label{fig:width-recovery}
    \end{figure}

MEME and QD both fail in the same place. Across the full length range neither ranks HNF4A above HNF4G, or JUND above its AP-1 relatives. Every MEME run has HNF4G, RXRB, JUN, FOSL2, or ATF3 as its top match, and the same relatives rank first in the archives in all but 0--7\% of seeds. The gap between any-hit and best-hit comes from near-identical reference motifs, which sequence-only searches will be unable to separate. The performance of the two methods diverge once length is constrained. Recovery within each fixed window is shown in Figure~\ref{fig:width-recovery}, and its sensitivity to the deduplication threshold in Table~\ref{tab:sens}. Across the five windows pooled, the archives recover RXRA in every seed and YY1 in 87--97\% of seeds, MEME loses YY1 in four windows and RXRA in one, and only the archives ever rank HNF4A first, in 30--43\% of seeds (Section~\ref{sec:discussion}). 

\subsection{Qualitative Results}
\label{sec:qualitative-results}
\begin{figure}[t]
    \centering
    \includegraphics[width=.85\linewidth]{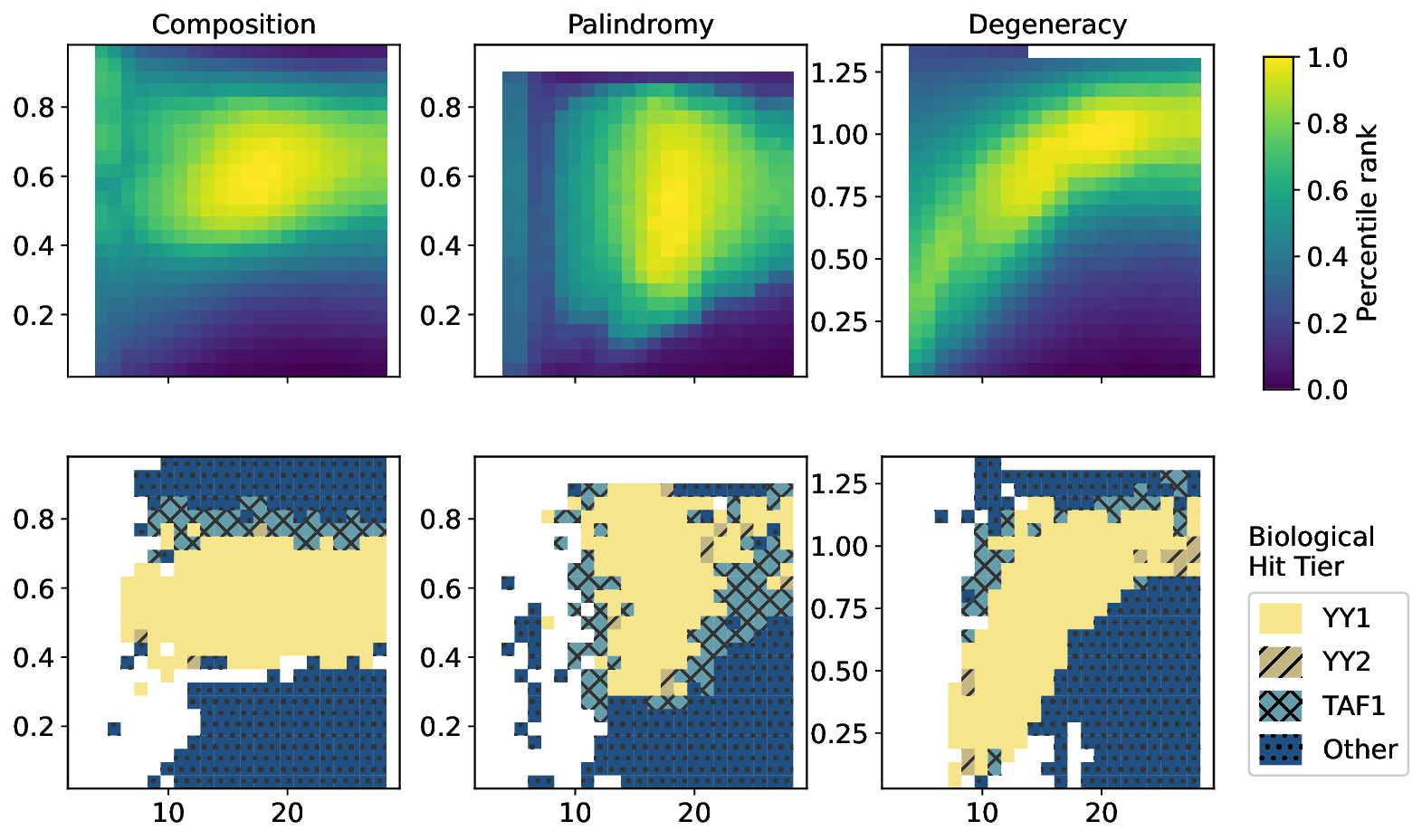}
    \caption{Top row: percentile rank of fitness per archive cell. Bottom row: biological hit tier per cell (TomTom match to YY1, YY2, TAF1, or other). Horizontal axes are effective motif length (bp); vertical axes are the respective descriptor value for each column (Composition, Palindromy, Degeneracy), each on its own scale.}
    \label{fig:composite}
    \end{figure}

The HNF4A result appears to be due to a length effect, visible in the HNF4A best-hit panel of Figure~\ref{fig:width-recovery}. Among archive cores that carry the HNF4A signal, HNF4G is the top match at every width, except in the 6--10 bp window, where HNF4A takes the top position in 31\% of cores and at least once in 32\% of seed archives. This falls to 0--2\% of lengths above 10 bp. When restricted to 6--10 bp, MEME finds a motif over the same element (BCAAAGKYCA), but HNF4G still ranks first, as it does in every other MEME window except 11--15 bp, where RXRB is first. Section~\ref{sec:discussion} examines what distinguishes the archive's short cores. The archives also show the competition directly. In YY1 the fitness peak in each archive aligns with the target-hit region, and the paralog YY2 and the partner TAF1 occupy the edge of that region in all descriptor pairs (Figure~\ref{fig:composite}). Here, the archive has an advantage over a ranked list because it keeps the target and its competitors side by side and shows how they separate along a structural axis. 

\section{Discussion}
\label{sec:discussion}
We compared MAP-Elites with a matched elitist baseline that differs only in how survivors are chosen. Although fitness is an essential component of motif discovery and locates where the target signal is, diversity preservation adds a further axis that fitness alone does not provide. The baseline usually converges onto a lower fitness peak than MAP-Elites; even in the runs where it reaches the same fitness, it collapses onto a single motif and fails to rank the target on three of five datasets. What separates the methods is whether the population keeps more than one peak.

Several distinct motifs can explain the same peaks equally well, so the single best match does not always identify the correct protein. A top-1 criterion therefore misstates success by an amount that depends on the target. For example, MEME's RXRA motif on the full length run has 32 significant matches, with RXRA fourth behind RARB, RXRB, and PPARA, so a best-hit criterion scores it as a failure. The same holds for MAP-Elites, where the highest-fitness YY1 motif in one archive (seed 1001) has 64 significant matches and YY1 sits at position 41. In discovery, the target motif is not known in advance. We recommend reporting every significant database match for each motif a search returns, along with how deep each candidate protein sits in the match list. When the target is known, as in our benchmark, both criteria should be reported together with the proteins ranked above the target. 

We found that MAP-Elites matches MEME's recovery on four datasets, underperforms on JUND, and is more robust to length, a parameter required by MEME that can strongly influence what it returns (Figure~\ref{fig:width-recovery}). Across fixed windows, the archives recover YY1 in 87--97\% of seeds and RXRA in every seed. In contrast, MEME fails to identify YY1 in four of five windows and RXRA at 21--25 bp. It keeps RXRA among its five closest matches in only one window, and drops its AP-1 motif from first at 6--10 bp to eighth at 26--30 bp. This holds across deduplication thresholds as any-hit rates move by at most one seed between 0.6 and 0.9, and HNF4A best-hit over pooled windows stays at 23--53\% of seeds against 0\% for MEME (Table~\ref{tab:sens}). We found that deduplication only changes magnitude, not direction, since counting near-copies raises that rate to 83--100\%. At the 0.75 similarity threshold an archive holds 2.0--9.7 distinct cores, against 5--9 for one MEME run (Table~\ref{tab:sens}), so its diversity is mostly behavioral, meaning spread along the descriptor axes of length, GC content, symmetry, or entropy, and not a breadth of different motifs.

\begin{table}[t]
    \centering
    \caption{Sensitivity to the core-similarity threshold. Distinct cores are means per archive at 6--30 bp (range over datasets) or per MEME run; HNF4A best-hit is the \% of seeds over pooled windows (Composition/Palindromy/Degeneracy).}\label{tab:sens}
    \setlength{\tabcolsep}{4pt}
    \scriptsize
    \begin{tabular}{lccccc}
    \toprule
    Threshold & Comp. cores & Pal. cores & Deg. cores & MEME cores & HNF4A best-hit\\
    \midrule
    0.60 & 3.2--4.4 & 1.6--2.6 & 1.6--3.8 & 4--6 & 43/37/23\\
    0.75 & 4.9--7.0 & 2.0--4.0 & 6.8--9.7 & 5--9 & 43/37/30\\
    0.90 & 11.7--16.3 & 3.3--9.6 & 21.7--27.0 & 9--10 & 50/47/53\\
    \bottomrule
    \end{tabular}
    \end{table}

We found that at short lengths the archive resolves the HNF4A motif. The HNF4A binding site exists in two near-identical variants, one bound only by HNF4 proteins and one shared with other nuclear receptors, which differ at two positions \cite{Fang2012}. Of 480 archive motifs carrying the target, 98\% match the HNF4-specific variant more closely, whereas MEME's motifs match both variants almost equally, because its statistical model averages the two versions. This is the familiar contrast between niching and convergence to an averaged optimum.

A few limitations bound these conclusions. First, we used one QD algorithm, so the evidence concerns diversity preservation as isolated by one matched baseline. Other techniques such as crowding, fitness sharing, novelty search, and other QD algorithms are untested, so we cannot say MAP-Elites is the best way to preserve diversity. Second, the datasets cover one tissue, five TFs, and one organism, so the discrimination difficulty we found may not generalize. Third, HOCOMOCO is partly derived from ChIP-seq data that may include the ENCODE experiments we analyze, so the validation may not be fully independent of the input. Fourth, the comparison is limited to a single classical tool in one mode, MEME's classic objective with default search settings and a single deterministic run per condition, with no discriminative or deep-learning tool, so we make no claim about where QD stands against the wider field. Fifth, the method does not reliably separate motif families in the more challenging settings. For the RXRA and HNF4A datasets, the competitors form a tight nuclear-receptor cluster, and for JUND the AP-1 relatives are close competitors and recovery is least reliable. In these cases, the archive exposes the presence of competition but does not resolve it.

\section{Conclusion}
\label{sec:conclusion}
We applied QD search to motif discovery with a variable-length genotype and validated every candidate against a full reference database and against MEME under identical inputs. Across five liver ChIP-seq datasets, a matched baseline without diversity preservation collapses onto single motifs and fails to rank the target first on three datasets, while MAP-Elites recovers the target signal as reliably as MEME on four datasets and more robustly across motif length. All methods show the same gap between recovering the sequence signal and identifying the specific TF, which leaves sequence annotation as an open problem. We argue that QD search is a feasible and informative approach to motif discovery. Future work should compare diversity mechanisms under one budget, design descriptors that target the features separating paralogs, and extend the benchmark to other tissues, organisms, and tools. 

\begin{credits}

\subsubsection{Statement of contributions.} A.M. designed the method, ran the experiments, analyzed results, and wrote the first draft. M.L.B. supervised the work and revised the paper. All authors approved the final version and take full responsibility for its content.

\subsubsection{Usage of AI tools.} GitHub Copilot was used to help write and debug the plotting scripts and the CSV file output utility scripts. GitHub Copilot was also used to help debug core components of the method when needed. Claude (Sonnet 5.5, Anthropic) was used to help format LaTeX tables and figures. All AI-generated code was checked and edited by the authors. No AI tool was used to generate results, analyze data, or produce references.

\subsubsection{\discintname}
The authors do not declare any conflict of interest.
\end{credits}

\bibliographystyle{splncs04}
\bibliography{references}

@inproceedings{medina_motif_2026,
  address = {Centro Internacional de Convenciones CIC-ANDE San Jose Costa Rica},
  title = {Motif {Diversity} in {Human} {Liver} {ChIP}-seq {Data} {Using} {MAP}-{Elites}},
  isbn = {979-8-4007-2488-6},
  doi = {10.1145/3795101.3805456},
  language = {en},
  urldate = {2026-08-14},
  booktitle = {Proceedings of the {Genetic} and {Evolutionary} {Computation} {Conference} {Companion}},
  publisher = {ACM},
  author = {Medina, Alejandro and Benton, Mary Lauren},
  month = jul,
  year = {2026},
  pages = {637--640},
}

@article{the_encode_project_consortium_integrated_2012,
  title = {An integrated encyclopedia of {DNA} elements in the human genome},
  volume = {489},
  copyright = {https://creativecommons.org/licenses/by-nc-sa/3.0/},
  issn = {0028-0836, 1476-4687},
  doi = {10.1038/nature11247},
  language = {en},
  number = {7414},
  urldate = {2026-06-12},
  journal = {Nature},
  author = {{The ENCODE Project Consortium}},
  month = sep,
  year = {2012},
  pages = {57--74},
}

@article{chalumeau2024qdax,
    title={Qdax: A library for quality-diversity and population-based algorithms with hardware acceleration},
    author={Chalumeau, Felix and Lim, Bryan and Boige, Raphael and Allard, Maxime and Grillotti, Luca and Flageat, Manon and Mac{\'e}, Valentin and Richard, Guillaume and Flajolet, Arthur and Pierrot, Thomas and others},
    journal={Journal of Machine Learning Research},
    volume={25},
    number={108},
    pages={1--16},
    year={2024}
}

@misc{mouret_illuminating_2015,
  title = {Illuminating search spaces by mapping elites},
  doi = {10.48550/arXiv.1504.04909},
  urldate = {2026-06-12},
  publisher = {arXiv},
  author = {Mouret, Jean-Baptiste and Clune, Jeff},
  month = apr,
  year = {2015},
  note = {arXiv:1504.04909 [cs.AI]},
}

@article{stormo_dna_2000,
  title = {{DNA} binding sites: representation and discovery},
  volume = {16},
  issn = {1367-4811, 1367-4803},
  shorttitle = {{DNA} binding sites},
  doi = {10.1093/bioinformatics/16.1.16},
  language = {en},
  number = {1},
  urldate = {2026-06-12},
  journal = {Bioinformatics},
  author = {Stormo, Gary D.},
  month = jan,
  year = {2000},
  pages = {16--23},
}

@inproceedings{che_mdga_2005,
  address = {Washington DC USA},
  title = {{MDGA}: motif discovery using a genetic algorithm},
  isbn = {978-1-59593-010-1},
  shorttitle = {{MDGA}},
  doi = {10.1145/1068009.1068080},
  language = {en},
  urldate = {2026-06-12},
  booktitle = {Proceedings of the 7th annual conference on {Genetic} and evolutionary computation},
  publisher = {ACM},
  author = {Che, Dongsheng and Song, Yinglei and Rasheed, Khaled},
  month = jun,
  year = {2005},
  pages = {447--452},
}

@article{gupta_quantifying_2007,
  title = {Quantifying similarity between motifs},
  volume = {8},
  issn = {1474-760X},
  doi = {10.1186/gb-2007-8-2-r24},
  language = {en},
  number = {2},
  urldate = {2026-06-12},
  journal = {Genome Biology},
  author = {Gupta, Shobhit and Stamatoyannopoulos, John A and Bailey, Timothy L and Noble, William Stafford},
  month = feb,
  year = {2007},
  pages = {R24},
}

@article{bailey_meme_2015,
  title = {The {MEME} {Suite}},
  volume = {43},
  copyright = {http://creativecommons.org/licenses/by/4.0/},
  issn = {0305-1048, 1362-4962},
  doi = {10.1093/nar/gkv416},
  language = {en},
  number = {W1},
  urldate = {2026-06-12},
  journal = {Nucleic Acids Research},
  author = {Bailey, Timothy L. and Johnson, James and Grant, Charles E. and Noble, William S.},
  month = jul,
  year = {2015},
  pages = {W39--W49},
}

@article{cully_robots_2015,
  title = {Robots that can adapt like animals},
  volume = {521},
  issn = {0028-0836, 1476-4687},
  doi = {10.1038/nature14422},
  language = {en},
  number = {7553},
  urldate = {2026-06-12},
  journal = {Nature},
  author = {Cully, Antoine and Clune, Jeff and Tarapore, Danesh and Mouret, Jean-Baptiste},
  month = may,
  year = {2015},
  pages = {503--507},
}

@inproceedings{vassiliades_mouret_isoline_2018,
  author = {Vassiliades, Vassiiis and Mouret, Jean-Baptiste},
  title = {Discovering the elite hypervolume by leveraging interspecies correlation},
  booktitle = {Proceedings of the Genetic and Evolutionary Computation Conference},
  year = {2018},
  isbn = {9781450356183},
  publisher = {Association for Computing Machinery},
  address = {New York, NY, USA},
  doi = {10.1145/3205455.3205602},
  pages = {149–156},
  numpages = {8},
  location = {Kyoto, Japan},
  series = {GECCO '18},
}

@ARTICLE{10.1109/TCBB.2022.3158471,
  author={Yu, Qiang and Zhang, Xiao and Hu, Yana and Chen, Shengpin and Yang, Liying},
  journal={IEEE/ACM Transactions on Computational Biology and Bioinformatics}, 
  title={A Method for {Predicting} {DNA} {Motif} {Length} {Based} {On} {Deep} {Learning}}, 
  year={2023},
  volume={20},
  number={1},
  pages={61-73},
  doi={10.1109/TCBB.2022.3158471}}

@article{Vorontsov2023,
  title = {{HOCOMOCO} in 2024: a rebuild of the curated collection of binding models for human and mouse transcription factors},
  volume = {52},
  issn = {1362-4962},
  doi = {10.1093/nar/gkad1077},
  number = {D1},
  journal = {Nucleic Acids Research},
  publisher = {Oxford University Press (OUP)},
  author = {Vorontsov, Ilya E and Eliseeva, Irina A and Zinkevich, Arsenii and Nikonov, Mikhail and Abramov, Sergey and Boytsov, Alexandr and Kamenets, Vasily and Kasianova, Alexandra and Kolmykov, Semyon and Yevshin, Ivan S and Favorov, Alexander and Medvedeva, Yulia A and Jolma, Arttu and Kolpakov, Fedor and Makeev, Vsevolod J and Kulakovskiy, Ivan V},
  year = {2023},
  month = Nov,
  pages = {D154–D163},
}

@article{Alipanahi2015,
  title = {Predicting the sequence specificities of {DNA}- and {RNA}-binding proteins by deep learning},
  volume = {33},
  issn = {1546-1696},
  doi = {10.1038/nbt.3300},
  number = {8},
  journal = {Nature Biotechnology},
  publisher = {Springer Science and Business Media LLC},
  author = {Alipanahi, Babak and Delong, Andrew and Weirauch, Matthew T and Frey, Brendan J},
  year = {2015},
  month = Jul,
  pages = {831–838},
}

@article{Bailey2021,
  title = {{STREME}: accurate and versatile sequence motif discovery},
  volume = {37},
  issn = {1367-4811},
  doi = {10.1093/bioinformatics/btab203},
  number = {18},
  journal = {Bioinformatics},
  publisher = {Oxford University Press (OUP)},
  author = {Bailey, Timothy L},
  editor = {Birol, Inanc},
  year = {2021},
  month = Mar,
  pages = {2834–2840},
}

@inproceedings{Bailey1994,
  author    = {Bailey, Timothy L. and Elkan, Charles},
  title     = {Fitting a {Mixture} {Model} by {Expectation} {Maximization} to {Discover} {Motifs} in {Biopolymers}},
  booktitle = {Proceedings of the Second International Conference on Intelligent Systems for Molecular Biology (ISMB)},
  volume    = {2},
  pages     = {28--36},
  publisher = {AAAI Press},
  year      = {1994}
}

@inproceedings{boige_gradient-informed_2023,
  address = {Lisbon Portugal},
  title = {Gradient-{Informed} {Quality} {Diversity} for the {Illumination} of {Discrete} {Spaces}},
  isbn = {979-8-4007-0119-1},
  doi = {10.1145/3583131.3590407},
  language = {en},
  urldate = {2026-06-12},
  booktitle = {Proceedings of the {Genetic} and {Evolutionary} {Computation} {Conference}},
  publisher = {ACM},
  author = {Boige, Raphaël and Richard, Guillaume and Dona, Jérémie and Pierrot, Thomas and Cully, Antoine},
  month = jul,
  year = {2023},
  pages = {119--128},
}

@inproceedings{Sambhe2021,
  title = {Multi-objective {Classification} and {Feature} {Selection} of {Covid}-19 {Proteins} {Sequences} using {NSGA}-II and {MAP}-{Elites}},
  doi = {10.5220/0010388512411248},
  booktitle = {Proceedings of the 13th International Conference on Agents and Artificial Intelligence},
  publisher = {SCITEPRESS - Science and Technology Publications},
  author = {Sambhe, Vijay and Rajesh, Shanmukha and Naredo, Enrique and Dias, Douglas and Kshirsagar, Meghana and Ryan, Conor},
  year = {2021},
  pages = {1241–1248},
}

@article{pugh_quality_2016,
  title = {Quality {Diversity}: {A} {New} {Frontier} for {Evolutionary} {Computation}},
  volume = {3},
  issn = {2296-9144},
  shorttitle = {Quality {Diversity}},
  doi = {10.3389/frobt.2016.00040},
  urldate = {2026-06-12},
  journal = {Frontiers in Robotics and AI},
  author = {Pugh, Justin K. and Soros, Lisa B. and Stanley, Kenneth O.},
  month = jul,
  year = {2016},
}

@article{Johnson2007,
  title = {{Genome-Wide Mapping of in Vivo Protein-DNA Interactions}},
  volume = {316},
  issn = {1095-9203},
  doi = {10.1126/science.1141319},
  number = {5830},
  journal = {Science},
  publisher = {American Association for the Advancement of Science (AAAS)},
  author = {Johnson, David S. and Mortazavi, Ali and Myers, Richard M. and Wold, Barbara},
  year = {2007},
  month = June,
  pages = {1497–1502},
}

@article{Weirauch2014,
  title = {{Determination and Inference of Eukaryotic Transcription Factor Sequence Specificity}},
  volume = {158},
  issn = {0092-8674},
  doi = {10.1016/j.cell.2014.08.009},
  number = {6},
  journal = {Cell},
  publisher = {Elsevier BV},
  author = {Weirauch, Matthew T. and Yang, Ally and Albu, Mihai and Cote, Atina G. and Montenegro-Montero, Alejandro and Drewe, Philipp and Najafabadi, Hamed S. and Lambert, Samuel A. and Mann, Ishminder and Cook, Kate and Zheng, Hong and Goity, Alejandra and van Bakel, Harm and Lozano, Jean-Claude and Galli, Mary and Lewsey, Mathew G. and Huang, Eryong and Mukherjee, Tuhin and Chen, Xiaoting and Reece-Hoyes, John S. and Govindarajan, Sridhar and Shaulsky, Gad and Walhout, Albertha J.M. and Bouget, François-Yves and Ratsch, Gunnar and Larrondo, Luis F. and Ecker, Joseph R. and Hughes, Timothy R.},
  year = {2014},
  month = Sept,
  pages = {1431–1443},
}

@article{Pabo1992,
  title = {{TRANSCRIPTION FACTORS: Structural Families and Principles of DNA Recognition}},
  volume = {61},
  issn = {1545-4509},
  doi = {10.1146/annurev.bi.61.070192.005201},
  number = {1},
  journal = {Annual Review of Biochemistry},
  publisher = {Annual Reviews},
  author = {Pabo, Carl O. and Sauer, Robert T.},
  year = {1992},
  month = June,
  pages = {1053–1095},
}

@article{Fang2012,
  title = {Identification of a binding motif specific to {HNF4} by comparative analysis of multiple nuclear receptors},
  volume = {40},
  issn = {0305-1048},
  doi = {10.1093/nar/gks190},
  number = {12},
  journal = {Nucleic Acids Research},
  publisher = {Oxford University Press (OUP)},
  author = {Fang, Bin and Mane-Padros, Daniel and Bolotin, Eugene and Jiang, Tao and Sladek, Frances M.},
  year = {2012},
  month = Mar,
  pages = {5343–5356},
}

@article{Xi2007,
  title = {Analysis of overrepresented motifs in human core promoters reveals dual regulatory roles of {YY1}},
  volume = {17},
  issn = {1088-9051},
  doi = {10.1101/gr.5754707},
  number = {6},
  journal = {Genome Research},
  publisher = {Cold Spring Harbor Laboratory},
  author = {Xi, Hualin and Yu, Yong and Fu, Yutao and Foley, Jonathan and Halees, Anason and Weng, Zhiping},
  year = {2007},
  month = June,
  pages = {798–806},
}

@article{Fonseca2019,
  title = {Diverse motif ensembles specify non-redundant {DNA} binding activities of {AP-1} family members in macrophages},
  volume = {10},
  issn = {2041-1723},
  doi = {10.1038/s41467-018-08236-0},
  number = {1},
  journal = {Nature Communications},
  publisher = {Springer Science and Business Media LLC},
  author = {Fonseca, Gregory J. and Tao, Jenhan and Westin, Emma M. and Duttke, Sascha H. and Spann, Nathanael J. and Strid, Tobias and Shen, Zeyang and Stender, Joshua D. and Sakai, Mashito and Link, Verena M. and Benner, Christopher and Glass, Christopher K.},
  year = {2019},
  month = Jan,
}

@article{Aptekmann2022,
  title = {Transcription factor specificity limits the number of {DNA}-binding motifs},
  volume = {17},
  issn = {1932-6203},
  doi = {10.1371/journal.pone.0263307},
  number = {1},
  journal = {PLOS ONE},
  publisher = {Public Library of Science (PLoS)},
  author = {Aptekmann, Ariel A. and Bulavka, Denys and Nadra, Alejandro D. and Sánchez, Ignacio E.},
  editor = {Huang, Chun-Hsi},
  year = {2022},
  month = Jan,
  pages = {e0263307},
}

@article{heinz_simple_2010,
  title = {Simple {Combinations} of {Lineage}-{Determining} {Transcription} {Factors} {Prime} cis-{Regulatory} {Elements} {Required} for {Macrophage} and {B} {Cell} {Identities}},
  volume = {38},
  copyright = {https://www.elsevier.com/tdm/userlicense/1.0/},
  issn = {10972765},
  doi = {10.1016/j.molcel.2010.05.004},
  language = {en},
  number = {4},
  urldate = {2026-06-12},
  journal = {Molecular Cell},
  author = {Heinz, Sven and Benner, Christopher and Spann, Nathanael and Bertolino, Eric and Lin, Yin C. and Laslo, Peter and Cheng, Jason X. and Murre, Cornelis and Singh, Harinder and Glass, Christopher K.},
  month = may,
  year = {2010},
  pages = {576--589},
}

@article{Vijayvargiya2011,
  title = {A niched {P}areto genetic algorithm for finding variable length regulatory motifs in {DNA} sequences},
  volume = {2},
  issn = {2190-5738},
  doi = {10.1007/s13205-011-0040-6},
  number = {2},
  journal = {3 Biotech},
  publisher = {Springer Science and Business Media LLC},
  author = {Vijayvargiya, Shripal and Shukla, Pratyoosh},
  year = {2011},
  month = Dec,
  pages = {141–148},
}

@article{Gonzlezlvarez2014,
  author  = {Gonz{\'a}lez-{\'A}lvarez, David L. and Vega-Rodr{\'i}guez, Miguel A. and Rubio-Largo, {\'A}lvaro},
  title   = {Multiobjective Optimization Algorithms for Motif Discovery in {DNA} Sequences},
  journal = {Genetic Programming and Evolvable Machines},
  year    = {2015},
  volume  = {16},
  number  = {2},
  pages   = {167--209},
  doi     = {10.1007/s10710-014-9232-2}
}

@article{Pugacheva2015,
  title = {Comparative analyses of {CTCF} and {BORIS} occupancies uncover two distinct classes of {CTCF} binding genomic regions},
  volume = {16},
  issn = {1474-760X},
  doi = {10.1186/s13059-015-0736-8},
  number = {1},
  journal = {Genome Biology},
  publisher = {Springer Science and Business Media LLC},
  author = {Pugacheva, Elena M. and Rivero-Hinojosa, Samuel and Espinoza, Celso A. and Méndez-Catalá, Claudia Fabiola and Kang, Sungyun and Suzuki, Teruhiko and Kosaka-Suzuki, Natsuki and Robinson, Susan and Nagarajan, Vijayaraj and Ye, Zhen and Boukaba, Abdelhalim and Rasko, John E. J. and Strunnikov, Alexander V. and Loukinov, Dmitri and Ren, Bing and Lobanenkov, Victor V.},
  year = {2015},
  month = Aug,
}

@article{Li2017,
  title = {Seeking {M}ultiple {S}olutions: {A}n {Updated Survey on Niching Methods and Their Applications}},
  volume = {21},
  issn = {1941-0026},
  doi = {10.1109/tevc.2016.2638437},
  number = {4},
  journal = {IEEE Transactions on Evolutionary Computation},
  publisher = {Institute of Electrical and Electronics Engineers (IEEE)},
  author = {Li, Xiaodong and Epitropakis, Michael G. and Deb, Kalyanmoy and Engelbrecht, Andries},
  year = {2017},
  month = Aug,
  pages = {518–538},
}

@article{Kheradpour2013,
  title = {{Systematic discovery and characterization of regulatory motifs in {ENCODE} {TF} binding experiments}},
  volume = {42},
  issn = {1362-4962},
  doi = {10.1093/nar/gkt1249},
  number = {5},
  journal = {Nucleic Acids Research},
  publisher = {Oxford University Press (OUP)},
  author = {Kheradpour, Pouya and Kellis, Manolis},
  year = {2013},
  month = Dec,
  pages = {2976–2987},
}

@inproceedings{Kent2022,
  series = {GECCO ’22},
  title = {A discretization-free metric for assessing quality diversity algorithms},
  doi = {10.1145/3520304.3534018},
  booktitle = {Proceedings of the Genetic and Evolutionary Computation Conference Companion},
  publisher = {ACM},
  author = {Kent, Paul and Branke, Juergen and Gaier, Adam and Mouret, Jean-Baptiste},
  year = {2022},
  month = July,
  pages = {2131–2135},
  collection = {GECCO ’22},
}

@article{Vierstra2020,
  title = {Global reference mapping of human transcription factor footprints},
  volume = {583},
  issn = {1476-4687},
  doi = {10.1038/s41586-020-2528-x},
  number = {7818},
  journal = {Nature},
  publisher = {Springer Science and Business Media LLC},
  author = {Vierstra, Jeff and Lazar, John and Sandstrom, Richard and Halow, Jessica and Lee, Kristen and Bates, Daniel and Diegel, Morgan and Dunn, Douglas and Neri, Fidencio and Haugen, Eric and Rynes, Eric and Reynolds, Alex and Nelson, Jemma and Johnson, Audra and Frerker, Mark and Buckley, Michael and Kaul, Rajinder and Meuleman, Wouter and Stamatoyannopoulos, John A.},
  year = {2020},
  month = July,
  pages = {729–736},
}

@article{Beyer2002,
  title = {Evolution strategies – {A} comprehensive introduction},
  volume = {1},
  issn = {1572-9796},
  doi = {10.1023/a:1015059928466},
  number = {1},
  journal = {Natural Computing},
  publisher = {Springer Science and Business Media LLC},
  author = {Beyer, Hans-Georg and Schwefel, Hans-Paul},
  year = {2002},
  month = Mar,
  pages = {3–52},
}

\end{document}